\documentclass[10pt,twocolumn,letterpaper]{article}

\usepackage{cvpr} 

\definecolor{cvprblue}{rgb}{0.21,0.49,0.74}
\usepackage[pagebackref,breaklinks,colorlinks,allcolors=cvprblue]{hyperref}
\usepackage{multirow}
\usepackage{pifont}
\usepackage{hhline}
\usepackage[percent]{overpic}

\def\paperID{37330}
\def\confName{CVPR}
\def\confYear{2026}
\definecolor{customPurple}{RGB}{214, 216, 235}

\title{Efficiency Follows Global-Local Decoupling}

\author{
\fontsize{11.2pt}{0pt}\selectfont {Zhenyu Yang$^{1,3}$,
Gensheng Pei$^{2*}$,
Tao Chen$^{1,3}$,
Yichao Zhou$^{1}$,
Tianfei Zhou$^{4}$,
Yazhou Yao$^{1,3*}$,
Fumin Shen$^{5}$}\\
\small\textsuperscript{\rm 1}Nanjing University of Science and Technology,\quad
\small\textsuperscript{\rm 2}Department of Electrical and Computer Engineering, Sungkyunkwan University\\
\small$^{3}$State Key Laboratory of Intelligent Manufacturing of Advanced Construction Machinery\\
\small\textsuperscript{\rm 4}Beijing Institute of Technology\qquad
\small\textsuperscript{\rm 5}University of Electronic Science and Technology of China\\
\small\url{https://github.com/NUST-Machine-Intelligence-Laboratory/ConvNeur}
}

\begin{document}
\maketitle
\begin{abstract}
Modern vision models must capture \textbf{image-level context} without sacrificing \textbf{local detail} while remaining computationally affordable. We revisit this tradeoff and advance a simple principle: \textbf{decouple} the roles of global reasoning and local representation. To operationalize this principle, we introduce \textbf{ConvNeur}, a two-branch architecture in which a lightweight neural memory branch aggregates global context on a compact set of tokens, and a locality-preserving branch extracts fine structure. A learned gate lets global cues modulate local features without entangling their objectives. This separation yields subquadratic scaling with image size, retains inductive priors associated with local processing, and reduces overhead relative to fully global attention. On standard classification, detection, and segmentation benchmarks, ConvNeur matches or surpasses comparable alternatives at similar or lower compute and offers favorable accuracy versus latency trade-offs at similar budgets. These results support the view that efficiency follows global-local decoupling.
\end{abstract}   

\section{Introduction}

Recent vision models are increasingly asked to do two things at once: reason over the entire image and preserve fine spatial detail. High-level tasks such as recognition and detection benefit from image-level context, since objects are often defined by scene layout, co-occurrence, or long-range shape cues. Meanwhile, many decisions occur at the pixel or patch level, where edges, textures, and thin structures must remain intact. This creates a fundamental tension: expanding the receptive field to ``\emph{see the whole}" typically relies on heavier global interactions, whereas keeping features locally faithful favors lightweight, inductive operations that stay close to the image grid. As resolutions rise, heads become denser, and deployment budgets stay tight, the question is not only how to model global information, but how to do so without eroding locality or blowing up compute.

\begin{figure}[t]
\centering
\includegraphics[width=1.0\linewidth]{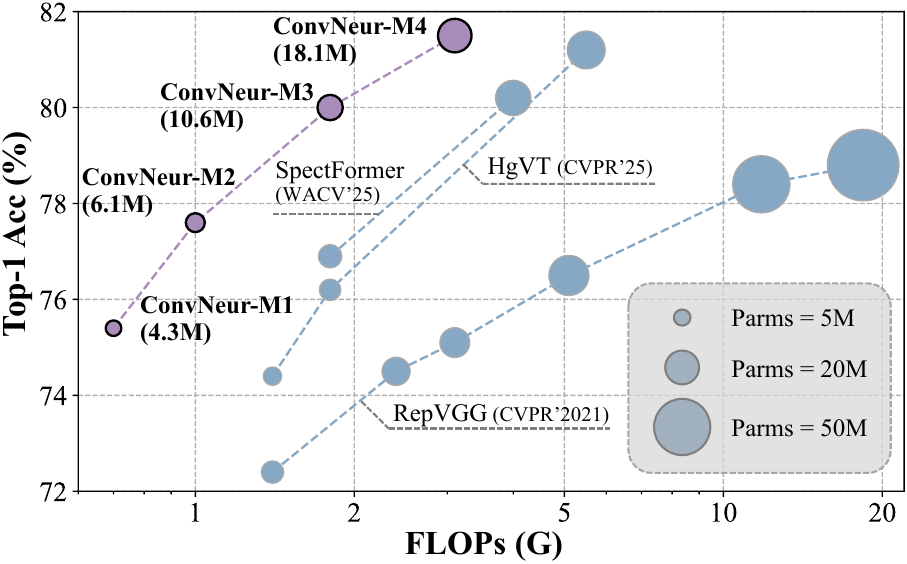}
\vspace{-0.6cm}
\caption{\textbf{Efficiency v.s. accuracy on the ImageNet-1K benchmark.} The proposed ConvNeur family establishes a new state-of-the-art frontier, delivering superior TOP-1 Acc while demanding significantly fewer FLOPs and parameters than existing methods.}
\label{fig:blob}
\vspace{-0.3cm}
\end{figure}

To address this “\emph{see the whole, keep the detail}” requirement, recent work has largely turned to Transformer-style global attention \cite{dosovitskiy2021vit, fuller2024lookhere, zhou2025unialign, pei2025seeing}. Its cost, however, still grows roughly \emph{quadratically} with resolution, which makes high-resolution or dense settings expensive. 
Windowed or sparse variants \cite{liu2021swin, hu2024lf} lower the cost but reintroduce locality and no longer view the full image. 
Convolutional models \cite{liu2022convnet, ding2022scaling, cai2024poly, pei2024videomac, cai2026beyond} remain efficient and preserve translation-friendly priors, yet their global context is limited or arrives late. 
Hybrid designs \cite{dai2021coatnet, zhu2024revisiting} try to combine the two, but global and local computation are usually carried out in the same feature space and at similar resolution, so the two roles remain coupled and the model effectively pays for both.

We observe that the main source of cost inflation is not global modeling itself, but the fact that global reasoning and local representation are often entangled in a single pathway. Once a feature stream is required to see the whole image, preserve fine-grained structure, and stay within a tight FLOP budget, width, spatial resolution, and interaction range start to compete and none of them is optimal. This suggests a different design principle: learn global and local separately, then let the global path modulate the local one instead of replacing it. A global path that only needs to guide can operate on compressed tokens and in chunked form, which keeps the computation subquadratic, while the local path can retain the inductive priors that make convolutions effective. In short, \emph{efficiency follows global-local decoupling}, not the other way around.

We realize this principle in \textbf{ConvNeur}, a two-branch vision architecture. The first branch is a locality-preserving path that follows the convolutional paradigm and focuses on edges, textures, and small structures and does not attempt to solve global reasoning. In parallel, a global branch bottlenecks the features to a smaller channel dimension, chunks the features into a sequence, and feeds each chunk into a neural memory module that aggregates image-level context in a blockwise manner. The chunkwise readouts are reassembled to the spatial layout and lifted to the original channel size to produce a global context map. A learned gate uses this map to modulate the local features, so the global branch acts as guidance while the local branch keeps the image priors. 
This differs from typical attention or hybrid designs in three ways. First, the global branch is not a full-width attention layer but a neural memory operating on compressed and chunked tokens, which makes the cost grow \emph{subquadratically} with image size. Second, fusion is done by gated modulation rather than naive concatenation or addition, so global signals do not wash out local evidence. Third, because the two roles are structurally decoupled, we can budget and compress the global branch independently of the local one, which is exactly what high-resolution vision needs.
We evaluate ConvNeur on ImageNet-1K \cite{deng2009imagenet} classification, COCO 2017 \cite{lin2014microsoft} detection, and ADE20K \cite{zhou2017scene} segmentation. Under comparable FLOPs and parameters, our models surpass recent efficient vision backbones, and the same module improves downstream tasks, supporting our claim that decoupling global and local leads to better accuracy-efficiency trade-offs.
Our main contributions are summarized as follows:
\begin{itemize}
    \item We identify global-local decoupling as a simple but effective route to efficiency in modern vision models.
    \item We introduce ConvNeur, a two-branch network that performs chunked neural memory aggregation on a compact global branch and uses learned gating to modulate a locality-preserving convolutional branch.
    \item On classification, detection, and segmentation tasks, the proposed ConvNeur achieves better accuracy-efficiency trade-offs than comparable convolutional and attention-based alternatives under similar budgets.
\end{itemize}

\section{Related Works}

\textbf{Vision Transformers \& Efficient Global Attention}. Vision Transformers (ViT) \cite{dosovitskiy2021vit} treated an image as a sequence of patches and applied fully global self-attention as the core vision operator, and DeiT \cite{touvron2021training} showed that with a stronger training recipe this paradigm is practical without massive external data. The limitation is structural, since all-to-all attention grows roughly quadratically with the number of tokens. Swin Transformer \cite{liu2021swin, liu2022swin} controlled this cost by restricting attention to local windows and shifting them across layers so that information can still flow globally. Follow-up windowed or hierarchical variants such as HEAL-Swin \cite{carlsson2024heal} and SHViT \cite{yun2024shvit} further tailored this design for large fields of view or for better memory usage. Another line \cite{li2024attnzero, zhou2024adapt} improves efficiency by reducing the active token set and learning sparse or adaptive attention patterns instead of enlarging windows. More recently, linearized and state-space vision backbones, including Vision Mamba \cite{zhu2024vision, huang2024localmamba, liu2024vmamba, hatamizadeh2025mambavision} and Vision RWKV \cite{duan2024vision}, replace quadratic attention with recurrent \cite{peng2023rwkv, peng2024eagle, peng2025rwkv} or SSM-style updates \cite{mamba, mamba2} to obtain long-range dependencies at near-linear cost. Across these directions, global interactions are made cheaper by being more local, more sparse, or more sequential, yet global and local information are still propagated in the same feature stream. Our approach is complementary: we place global reasoning on a compressed, chunked path and let it modulate a separate locality-preserving path, removing the coupling rather than only approximating attention.

\noindent\textbf{Convolutional \& Hybrid Vision Backbones}. Early CNNs \cite{simonyan2014very, szegedy2015going, he2016deep} struggle to capture image-level context due to their inherently local receptive fields. ConvNeXt \cite{liu2022convnet, woo2023convnext} addressed this with depthwise convolutions, layer-wise normalization, and later GRN and improved scaling. Large kernel models such as RepLKNet \cite{ding2022scaling} further enlarged the receptive field to approximate global context, while LSNet \cite{wang2025lsnet} paired a large-context view with a detail-preserving path. In parallel, hybrid models like CoAtNet \cite{dai2021coatnet} and \cite{zhu2024revisiting} used convolutions for local structure and attention or token mixing for global interactions. These approaches improve recognition, but global and local computation are still carried out in the same feature space, so the two remain coupled. Our method keeps the “conv for local, another operator for global’’ idea, but explicitly separates the two paths and fuses them only through a learned gate, which allows the global branch to be budgeted independently.

\begin{figure*}[t]
  \centering
  \begin{overpic}[width=\linewidth,grid=false]{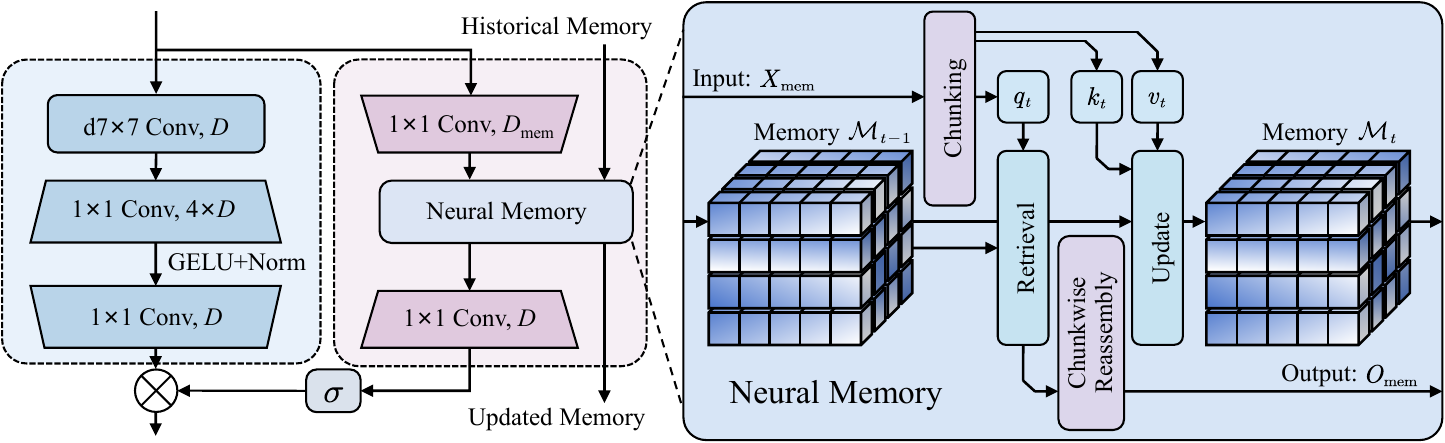}
    \put(12.2,24.7){\textbf{\S \ref{sec:3.2}}}
    \put(34,24.7){\textbf{\S \ref{sec:3.3}}}
  \end{overpic}
  \caption{\textbf{The overall architecture of ConvNeur.} On the left, a locality-preserving convolutional branch extracts fine-grained features. In parallel, a compressed global branch performs chunked neural memory aggregation, and produces a gating map to modulate the local features. On the right, we show the neural-memory module: each chunk is linearly mapped to $\{q_t,k_t,v_t\}$. $q_t$ reads from the previous memory state $\mathcal{M}_{t-1}$, while $k_t$ and $v_t$ compute a surprise loss to update the memory to $\mathcal{M}_t$.}
  \label{fig:convneur}
  \vspace{-0.1cm}
\end{figure*}

\noindent\textbf{Memory-Augmented Models}. Recent vision works \cite{agro2025mad, zhou2025splatmesh, yang2025changetitans} began to attach explicit memory to the network, especially for video and open-world settings. RMem \cite{zhou2024rmem} keeps a restricted memory bank and reads past object cues back into the current frame, while CUTIE \cite{cheng2024putting} performs object-level memory reading to restore instance consistency over time. CMeRT \cite{pang2025context} uses a context-enhanced encoder and a memory-refined decoder to exploit near-past features. Earlier memory-augmented models in vision, such as Non-local Networks \cite{wang2018non}, GCNet \cite{cao2019gcnet}, and CBAM \cite{woo2018cbam}, also aggregate global signals first and then modulate local features. Video memory networks like STM \cite{oh2019video} and AOT \cite{yang2021associating} keep a spatiotemporal memory that is read to guide the current prediction. Titans \cite{behrouz2024titans} goes one step further by organizing memory into chunks, using surprise-driven updates, and treating memory as a learnable component rather than a fixed repository. Our work follows the same spirit of separating fast locality-preserving computation from a slower context-accumulating component, but we do this spatially in an explicit two-branch architecture.

\section{Method}

\subsection{Overview Architecture of ConvNeur}

Given an intermediate feature map $X\in\mathbb{R}^{C\times H\times W}$, our objective is to model image-level context and local structure separately and to let the former modulate the latter. The guiding principle is that efficiency follows global-local decoupling: global reasoning and locality-preserving representation should not compete for width, spatial resolution, or interaction range inside the same pathway.

As shown in Fig.~\ref{fig:convneur}, we design ConvNeur as a two-branch architecture. The locality-preserving convolutional branch follows the modern convolutional paradigm \cite{liu2022convnet} and is responsible only for edges, textures, and other fine-scale patterns; it keeps the inductive priors of CNNs and does not attempt to solve global reasoning. In parallel, the compressed global-memory branch first projects $X$ to a lower-dimensional memory space, converts the spatial map to a sequence, partitions it into fixed-size chunks, and for each chunk performs a \textit{retrieval-update-reassemble} cycle that aggregates image-level context at subquadratic cost. This branch is inspired by recent neural memory \cite{behrouz2024titans} formulations, so we adopt the term \textit{neural memory} and the surprise-driven update mechanism, but we place it inside a spatially decoupled vision framework so that global context is produced on a lightweight path and only used to gate the convolutional features rather than replacing them.

\subsection{Locality-Preserving Convolutional Branch}\label{sec:3.2}
The first branch keeps feature processing local. It applies a depthwise convolution with a relatively large kernel to gather nearby context, followed by lightweight channel mixing and normalization. This path is meant to capture edges, textures and small objects and to retain the convolutional inductive bias. We denote its output by $F_\text{loc}$. Because global reasoning is handled elsewhere, this branch does not expand its interaction range and therefore avoids the width-resolution-context entanglement discussed above.

\subsection{Global Memory Branch}\label{sec:3.3}

The global branch is built around an optimization-style view. The outer part of the network learns how global context should be queried and how memory updates should be parametrized, while an inner, chunk-wise procedure actually updates the memory state for the current image. 
This yields a structure that is close to bilevel optimization \cite{finn2017model, franceschi2018bilevel, petrulionyte2024functional}, where slow variables specify the update rule and fast variables adapt to the current example, and also to online or block-coordinate methods that process one block at a time to control complexity. 
Processing the image in blocks is exactly what keeps the cost below quadratic.

\noindent\textbf{Bottlenecked tokenization}. We first apply a pointwise projection to map the feature map $X$ into a lower-dimensional memory space, obtaining $X_\text{mem}\in\mathbb{R}^{C_m\times H\times W}$, $C_m<C$. This gives the global branch a compressed representation on which reading and updating the memory is more efficient, while preserving the spatial layout for later reassembly.

\begin{table}[t]
\centering
\caption{\textbf{ConvNeur variants used in our experiments.} Depths and dims are listed from high to low resolution. Memory dimension is matched to the stage width, and the chunk size is shared.}
\vspace{-0.2cm}
\label{tab:convneur_variants}
\setlength{\tabcolsep}{3pt}
\resizebox{\columnwidth}{!}{%
\begin{tabular}{l|cccc|c}
\toprule
\textbf{Model} & \textbf{Depths} & \textbf{Dims} & \textbf{\begin{tabular}[c]{@{}c@{}}Memory\\ Dim $C_m$\end{tabular}} & \textbf{\begin{tabular}[c]{@{}c@{}}Chunk\\ Size $L$\end{tabular}} & \textbf{\begin{tabular}[c]{@{}c@{}}FLOPs\\ (G)\end{tabular}} \\
\midrule
ConvNeur-M1 & [2, 2, 6, 2] & [40, 80, 160, 320] & 80  & 196 & 0.71 \\
ConvNeur-M2 & [2, 2, 6, 2] & [48, 96, 192, 384] & 96  & 196 & 1.01 \\
ConvNeur-M3 & [2, 2, 6, 2] & [64, 128, 256, 512] & 128 & 196 & 1.77 \\
ConvNeur-M4 & [2, 2, 8, 2] & [80, 160, 320, 640] & 160 & 196 & 3.06 \\
\bottomrule
\end{tabular}}
\vspace{-0.2cm}
\end{table}

\noindent\textbf{Chunking}. We flatten $X_\text{mem}$ along the spatial dimensions to obtain $S\in\mathbb{R}^{(HW)\times C_m}$ and then partition this sequence into fixed-size chunks $\{S_t\}_{t=1}^{T}$, each of length $L$. Processing the image in chunks can be viewed as an online or block-coordinate pass over the spatial sequence. At each step the memory only receives a small subset of tokens, which prevents the global branch from operating on all tokens at once.

\noindent\textbf{Memory Retrieval \& Update}. For each chunk $S_t\in\mathbb{R}^{L\times C_m}$, we produce three sequences with linear projection:
\begin{equation}
    q_t=W_qS_t,\quad k_t=W_kS_t,\quad v_t=W_vS_t,
\end{equation}
where $\{W_q,W_k,W_v\}$ are learned together with the backbone and determine how the memory is queried and how it is updated.
We first use $q_t$ to read from the current memory state $\mathcal{M}_{t-1}$ and obtain a chunk-level global context $\hat{y}_t$. Given the current memory, this step selects the global information that is most useful for this part of the image:
\begin{equation}
    \hat{y}_t=\mathcal{M}_{t-1}(q_t).
\end{equation}
We then define for this chunk a reconstruction objective:
\begin{equation}
    \mathcal{L}_t=\|\mathcal{M}_{t-1}(k_t)-v_t\|_2^2.
\end{equation}
A set of learnable update generators converts this loss into an adaptive step size, a momentum coefficient, and a decay factor, and applies them to obtain the next memory state:
\begin{equation}
    \mathcal{M}_{t}=\mathcal{U}(\mathcal{M}_{t-1},\nabla_\mathcal{M}\mathcal{L}_t).
\end{equation}
Chunks that are harder to reconstruct produce larger updates, which follows the surprise-driven idea in recent neural memory work \cite{behrouz2024titans}. Because we repeat this \textit{Retrieval-Update} pattern for every chunk and for every head, different chunks can maintain partially independent fast weights and the memory behaves like a small ensemble.

\noindent\textbf{Chunkwise reassembly}. The retrieved contexts $\{\hat{y}_t\}_{t=1}^T$ are concatenated in the original token order to form a full sequence of length $HW$, which is then reshaped back to the spatial layout $O_\text{mem}\in\mathbb{R}^{C_m\times H\times W}$ to ensure that, although global modeling is carried out chunk by chunk, the final output is aligned with the original feature map $X_\text{mem}$.

\noindent\textbf{Global map}. After chunkwise reassembly, we restore the channel dimension with another pointwise convolution to obtain the global context map $G\in\mathbb{R}^{C\times H\times W}$. 

The total cost of the global branch can be written as:
\begin{equation}
    \mathcal{O}(CC_mHW)+\mathcal{O}(T\cdot\text{mem}(L,C_m)).
\end{equation}

Since the memory dimension $C_m$ and the chunk size $L$ are small and fixed in our models, the overall cost grows roughly linearly with the number of spatial locations.

\subsection{Gated Global-to-Local Fusion}

We pass the global context map through a sigmoid function ($\sigma$) to obtain a spatial gate $A=\sigma(G)$, and use it to modulate the local features pointwise:
\begin{equation}
    F_\text{out}=X+\mathtt{DropPath}(A\otimes F_\text{loc}).
\end{equation}
This can be viewed as amortized conditioning: the global branch first infers a context that is specific to the current image and then converts it into a mask that tells the local branch what to emphasize or suppress. The global signal therefore guides rather than overwrites the local representation, so convolutional priors are preserved. Compared with SE \cite{hu2018squeeze}, CBAM \cite{woo2018cbam} or non-local \cite{wang2018non}, which aggregate global information once and broadcast back, our gate is produced from an online-updated memory over spatial chunks, and thus reflects the current state of the global branch.

\subsection{Model Variants and Scaling}

To show that our design is not limited to a single width-depth configuration, we instantiate ConvNeur in several sizes. All variants share the same template: four stages with downsampling, a locality preserving branch in every block, and a global memory branch inserted at the start of each stage. As listed in Tab.~\ref{tab:convneur_variants}, we scale the models by widening the stage channels and proportionally increasing the memory dimension so that the global path stays lightweight relative to the local path. We keep the chunk size fixed, which makes the memory cost predictable across variants.

\begin{table}[]
\caption{\textbf{Comparison results of vision backbone performance on the ImageNet-1K \cite{deng2009imagenet} dataset pre-training.}}
\label{tab:imagenet}
\vspace{-0.2cm}
\setlength{\tabcolsep}{4pt}
\resizebox{\columnwidth}{!}{%
\begin{tabular}{lcccc}
\specialrule{1pt}{0pt}{0pt}
\rowcolor[HTML]{D6D8EB} 
\textbf{Method}                             & \textbf{Venue}        
& \textbf{\begin{tabular}[c]{@{}c@{}}Params\\ (M) $\downarrow$\end{tabular}}    
& \textbf{\begin{tabular}[c]{@{}c@{}}FLOPs\\ (G) $\downarrow$\end{tabular}}     
& \textbf{\begin{tabular}[c]{@{}c@{}}Top-1\\ Acc $\uparrow$\end{tabular}}     \\ \hline
\rowcolor[HTML]{F2F2F2} 
RepVGG-A0 \cite{ding2021repvgg}             & CVPR 2021    & 8.3           & 1.4          & 72.4          \\
\rowcolor[HTML]{F2F2F2} 
DeiT-T \cite{touvron2021training}           & ICML 2021    & 5.7           & 1.1          & 72.2          \\
\rowcolor[HTML]{F2F2F2} 
ConViT-Ti \cite{d2021convit}                & ICML 2021    & 5.7           & 1.0          & 73.1          \\
\rowcolor[HTML]{F2F2F2} 
CrossViT-Ti \cite{chen2021crossvit}         & ICCV 2021    & 6.9           & 1.6          & 73.4          \\
\rowcolor[HTML]{F2F2F2} 
PVT-T \cite{wang2021pyramid}                & ICCV 2021    & 13.2          & 1.9          & 75.1          \\
\rowcolor[HTML]{F2F2F2} 
GFNet-Ti \cite{rao2021global}               & NeurIPS 2021 & 7.0           & 1.3          & 74.6          \\
\rowcolor[HTML]{F2F2F2} 
PoolFormer-S7 \cite{yu2022metaformer}       & CVPR 2022    & 8.6           & 1.1          & 73.0          \\
\rowcolor[HTML]{F2F2F2} 
ViG-Ti \cite{han2022vision}                 & NeurIPS 2022 & 7.1           & 1.3          & 73.9          \\
\rowcolor[HTML]{F2F2F2} 
DependencyViT-Lite-T \cite{ding2023visual}  & CVPR 2023    & 6.2           & 0.8          & 73.7          \\
\rowcolor[HTML]{F2F2F2} 
ViHGNN-Ti \cite{han2023vision}              & ICCV 2023    & 8.2           & 1.8          & 74.3          \\
\rowcolor[HTML]{F2F2F2} 
MobileAtt-DeiT-T \cite{yao2024mobile}       & ICML 2024    & 5.7           & 1.2          & 73.3          \\
\rowcolor[HTML]{F2F2F2} 
SLAB-DeiT-T \cite{guo2024slab}              & ICML 2024    & 6.2           & 1.3          & 73.6          \\
\rowcolor[HTML]{F2F2F2} 
Agent-DeiT-T \cite{han2024agent}            & ECCV 2024    & 6.0           & 1.2          & 74.9          \\
\rowcolor[HTML]{F2F2F2} 
QuadMamba-Li \cite{xie2024quadmamba}        & NeurIPS 2024 & 5.4           & 0.8          & 74.2          \\
\rowcolor[HTML]{F2F2F2} 
KAT-T \cite{yang2025kolmogorov}             & ICLR 2025    & 5.7           & 1.1          & 74.6          \\
\rowcolor[HTML]{F2F2F2} 
VRWKV-T \cite{duan2024vision}               & ICLR 2025    & 6.2           & 1.2          & 75.1          \\
\rowcolor[HTML]{F2F2F2} 
HgVT-Mi \cite{fixelle2025hypergraph}        & CVPR 2025    & 5.8           & 1.4          & 74.4          \\
\rowcolor[HTML]{D6D8EB} 
\textbf{ConvNeur-M1(ours)}                  & -            & \textbf{4.3}  & \textbf{0.7} & \textbf{75.4} \\ \hline
\rowcolor[HTML]{F2F2F2} 
RepVGG-A2 \cite{ding2021repvgg}             & CVPR 2021    & 25.5          & 5.1          & 76.5          \\
\rowcolor[HTML]{F2F2F2} 
ConViT-Ti+ \cite{d2021convit}               & ICML 2021    & 10.0          & 2.0          & 76.7          \\
\rowcolor[HTML]{F2F2F2} 
CrossViT-9 \cite{chen2021crossvit}          & ICCV 2021    & 8.8           & 2.0          & 77.1          \\
\rowcolor[HTML]{F2F2F2} 
PoolFormer-S12 \cite{yu2022metaformer}      & CVPR 2022    & 12.0          & 1.8          & 77.2          \\
\rowcolor[HTML]{F2F2F2} 
EdgeViT-XS \cite{pan2022edgevits}           & ECCV 2022    & 6.7           & 1.1          & 77.5          \\
\rowcolor[HTML]{F2F2F2} 
RIFormer-S12 \cite{wang2023riformer}        & CVPR 2023    & 12.0          & 1.8          & 76.9          \\
\rowcolor[HTML]{F2F2F2} 
MobileOne-S2 \cite{vasu2023mobileone}       & CVPR 2023    & 7.8           & 1.3          & 77.4          \\
\rowcolor[HTML]{F2F2F2} 
SLAB-PVT-T \cite{guo2024slab}               & ICML 2024    & 13.4          & 1.9          & 76.0          \\
\rowcolor[HTML]{F2F2F2} 
Vim-Ti \cite{zhu2024vision}                 & ICML 2024    & 7.0           & 1.5          & 76.1          \\
\rowcolor[HTML]{F2F2F2} 
LocalVim-T \cite{huang2024localmamba}       & ECCV 2024    & 8.0           & 1.5          & 75.8          \\
\rowcolor[HTML]{F2F2F2} 
HgVT-Ti \cite{fixelle2025hypergraph}        & CVPR 2025    & 7.7           & 1.8          & 76.2          \\
\rowcolor[HTML]{F2F2F2} 
SpectFormer-T \cite{patro2025spectformer}   & WACV 2025    & 9.0           & 1.8          & 76.9          \\
\rowcolor[HTML]{D6D8EB} 
\textbf{ConvNeur-M2(ours)}                  & -            & \textbf{6.1}  & \textbf{1.0} & \textbf{77.6} \\ \hline
\rowcolor[HTML]{F2F2F2} 
RepVGG-B1g2 \cite{ding2021repvgg}           & CVPR 2021    & 41.4          & 8.8          & 77.8          \\
\rowcolor[HTML]{F2F2F2} 
DeiT-S  \cite{touvron2021training}          & ICML 2021    & 22.1          & 4.6          & 79.9          \\
\rowcolor[HTML]{F2F2F2} 
PVT-S \cite{wang2021pyramid}                & ICCV 2021    & 24.5          & 3.8          & 79.8          \\
\rowcolor[HTML]{F2F2F2} 
GFNet-XS \cite{rao2021global}               & NeurIPS 2021 & 16.0          & 2.9          & 78.6          \\
\rowcolor[HTML]{F2F2F2} 
FasterNet-T2 \cite{chen2023run}             & CVPR 2023    & 15.0          & 1.9          & 78.9          \\
\rowcolor[HTML]{F2F2F2} 
MobileOne-S4 \cite{vasu2023mobileone}       & CVPR 2023    & 14.8          & 3.0          & 79.4          \\
\rowcolor[HTML]{F2F2F2} 
Flatten-PVTv2-B1 \cite{han2023flatten}      & CVPR 2023    & 19.2          & 2.2          & 79.5          \\
\rowcolor[HTML]{F2F2F2} 
GC ViT-XXT \cite{hatamizadeh2023global}     & ICML 2023    & 12.0          & 2.1          & 79.9          \\
\rowcolor[HTML]{F2F2F2} 
Agent-PVT-T \cite{han2024agent}             & ECCV 2024    & 11.6          & 2.0          & 78.4          \\
\rowcolor[HTML]{D6D8EB} 
\textbf{ConvNeur-M3(ours)}                  & -            & \textbf{10.6} & \textbf{1.8} & \textbf{80.0} \\ \hline
\rowcolor[HTML]{F2F2F2} 
RepVGG-B3g4 \cite{ding2021repvgg}           & CVPR 2021    & 75.6          & 16.1         & 80.2          \\
\rowcolor[HTML]{F2F2F2} 
CrossViT-S \cite{chen2021crossvit}          & ICCV 2021    & 26.7          & 5.6          & 81.0          \\
\rowcolor[HTML]{F2F2F2} 
PVT-M \cite{wang2021pyramid}                & ICCV 2021    & 44.2          & 6.7          & 81.2          \\
\rowcolor[HTML]{F2F2F2} 
GFNet-B \cite{rao2021global}                & NeurIPS 2021 & 43.0          & 7.9          & 80.7          \\
\rowcolor[HTML]{F2F2F2} 
PoolFormer-S24 \cite{yu2022metaformer}      & CVPR 2022    & 21.0          & 3.4          & 80.3          \\
\rowcolor[HTML]{F2F2F2} 
ViG-S \cite{han2022vision}                  & NeurIPS 2022 & 22.7          & 4.5          & 80.4          \\
\rowcolor[HTML]{F2F2F2} 
RIFormer-S24 \cite{wang2023riformer}        & CVPR 2023    & 21.0          & 3.4          & 80.3          \\
\rowcolor[HTML]{F2F2F2} 
Vim-S \cite{zhu2024vision}                  & ICML 2024    & 26.0          & 5.3          & 80.5          \\
\rowcolor[HTML]{F2F2F2} 
LocalVim-S \cite{huang2024localmamba}       & ECCV 2024    & 28.0          & 4.8          & 81.0          \\
\rowcolor[HTML]{F2F2F2} 
VRWKV-S \cite{duan2024vision}               & ICLR 2025    & 23.8          & 4.6          & 80.1          \\
\rowcolor[HTML]{F2F2F2} 
KAT-S \cite{yang2025kolmogorov}             & ICLR 2025    & 22.1          & 4.4          & 81.2          \\
\rowcolor[HTML]{F2F2F2} 
HgVT-S \cite{fixelle2025hypergraph}         & CVPR 2025    & 22.9          & 5.5          & 81.2          \\
\rowcolor[HTML]{F2F2F2} 
SpectFormer-XS  \cite{patro2025spectformer} & WACV 2025    & 20.0          & 4.0          & 80.2          \\
\rowcolor[HTML]{D6D8EB} 
\textbf{ConvNeur-M4(ours)}                  & -            & \textbf{18.1} & \textbf{3.1} & \textbf{81.5} \\ 
\specialrule{1pt}{0pt}{0pt}
\end{tabular}}
\vspace{-0.45cm}
\end{table}

\section{Experiments}
We evaluate the proposed ConvNeur on public benchmarks for image classification \cite{deng2009imagenet}, object detection \cite{lin2014microsoft}, and semantic segmentation \cite{zhou2017scene} under standard protocols with matched compute and parameter budgets to rigorously and fairly assess effectiveness, scalability, and efficiency.

\subsection{Image Classification}\label{sec:cls} 

\textbf{Settings}. We train on ImageNet-1K with 224$\times$224 random crops for 300 epochs. The optimizer is AdamW \cite{loshchilov2017decoupled} with learning rate 4e-3 at a global batch size of 4096, epsilon 1e-8, weight decay 0.05 with cosine decay to a final value when specified, no gradient clipping, layer decay 1.0, and minimum learning rate 1e-6. The learning rate is warmed up for 20 epochs. When warmup steps are provided, the step schedule overrides the epoch count. Training uses 8 NVIDIA RTX 3090 GPUs. Data augmentation follows the widely used DeiT \cite{touvron2021training} style recipe: RandAugment, Mixup, CutMix, Random Erasing, and label smoothing.

\begin{table}[]
\caption{\textbf{Comparison results of object detection and instance segmentation performance on the COCO 2017 \cite{lin2014microsoft} validation set using Cascade Mask R-CNN \cite{cai2019cascade} under the 1x schedule.} All methods share the same detector, training recipe, and data pipeline.}
\label{tab:coco}
\vspace{-0.2cm}
\setlength{\tabcolsep}{7pt}
\resizebox{\columnwidth}{!}{%
\begin{tabular}{
>{\columncolor[HTML]{ECF4FF}}c
>{\columncolor[HTML]{EFEFEF}}l 
>{\columncolor[HTML]{EFEFEF}}c 
>{\columncolor[HTML]{EFEFEF}}c 
>{\columncolor[HTML]{EFEFEF}}c }
\specialrule{1pt}{0pt}{0pt}
\cellcolor[HTML]{D6D8EB}\textbf{Task} & \cellcolor[HTML]{D6D8EB}\textbf{Method} 
& \cellcolor[HTML]{D6D8EB}\textbf{$\text{AP}$$\uparrow$} 
& \cellcolor[HTML]{D6D8EB}\textbf{$\text{AP}_{50}$$\uparrow$} 
& \cellcolor[HTML]{D6D8EB}\textbf{$\text{AP}_{75}$$\uparrow$}                      \\\hline\hline
& ResNet50 \cite{he2016deep}                & 38.0 & 58.6          & 41.4          \\
& PVT-T \cite{wang2021pyramid}              & 36.7 & 59.2          & 39.3          \\
& PVTv2-B0 \cite{wang2022pvt}               & 38.2 & 60.5          & 40.7          \\
& BiViT \cite{he2023bivit}                  & 40.8 & 59.2          & 44.1          \\
& \cellcolor[HTML]{D6D8EB}\textbf{ConvNeur-M2 (ours)}
& \cellcolor[HTML]{D6D8EB}\textbf{41.2} 
& \cellcolor[HTML]{D6D8EB}\textbf{60.9}              
& \cellcolor[HTML]{D6D8EB}\textbf{44.5}                                            \\\hhline{~----}
& ResNet101 \cite{he2016deep}               & 40.4 & 61.1          & 44.2          \\
& ResNeXt101-32x4d \cite{xie2017aggregated} & 41.9 & 62.5          & 45.9          \\
& PVT-S \cite{wang2021pyramid}              & 40.4 & 62.9          & 43.8          \\
& PVTv2-B1 \cite{wang2022pvt}               & 41.8 & 64.3          & 45.9          \\
\multirow{-10}{*}{\begin{tabular}[c]{@{}c@{}}Object\\ Detection\end{tabular}} 
& \cellcolor[HTML]{D6D8EB}\textbf{ConvNeur-M3 (ours)}
& \cellcolor[HTML]{D6D8EB}\textbf{42.4}
& \cellcolor[HTML]{D6D8EB}\textbf{64.8}
& \cellcolor[HTML]{D6D8EB}\textbf{46.1}                                            \\\hline\hline
& ResNet50 \cite{he2016deep}                & 34.4 & 55.1          & 36.7          \\
& PVT-T \cite{wang2021pyramid}              & 35.1 & 56.7          & 37.3          \\
& PVTv2-B0 \cite{wang2022pvt}               & 36.2 & \textbf{57.8} & \textbf{38.6} \\
& BiViT \cite{he2023bivit}                  & 35.7 & 56.5          & 38.2          \\
& \cellcolor[HTML]{D6D8EB}\textbf{ConvNeur-M2 (ours)} 
& \cellcolor[HTML]{D6D8EB}\textbf{36.4} 
& \cellcolor[HTML]{D6D8EB}57.7 
& \cellcolor[HTML]{D6D8EB}\textbf{38.6}                                            \\\hhline{~----}
& ResNet101 \cite{he2016deep}               & 36.4 & 57.7          & 38.8          \\
& ResNeXt101-32x4d \cite{xie2017aggregated} & 37.5 & 59.4          & 40.2          \\
& PVT-S \cite{wang2021pyramid}              & 37.8 & 60.1          & 40.3          \\
& PVTv2-B1 \cite{wang2022pvt}               & 38.8 & \textbf{61.2} & 41.6          \\
\multirow{-10}{*}{\begin{tabular}[c]{@{}c@{}}Instance\\ Segmentation\end{tabular}} 
& \cellcolor[HTML]{D6D8EB}\textbf{ConvNeur-M3 (ours)}
& \cellcolor[HTML]{D6D8EB}\textbf{39.0}
& \cellcolor[HTML]{D6D8EB}61.0
& \cellcolor[HTML]{D6D8EB}\textbf{41.8}                                            \\
\specialrule{1pt}{0pt}{0pt}
\end{tabular}}
\vspace{-0.4cm}
\end{table}

\noindent\textbf{Results}. Tab.~\ref{tab:imagenet} shows the proposed ConvNeur advances the accuracy-efficiency frontier across budgets with only a few reference points needed. At the small end, M1 reaches 75.4 with 0.7 GFLOPs and 4.3M parameters, outperforming PVT-T \cite{wang2021pyramid} at 75.1 with 1.9 GFLOPs while using less than half the compute. In the mid range, M3 attains 80.0 with 1.8 GFLOPs and 10.6M parameters, matching or exceeding DeiT-S \cite{touvron2021training} at 79.9 with 4.6 GFLOPs and PVT-S at 79.8 with 3.8 GFLOPs at a much lower cost. At the high end, M4 delivers 81.5 with 3.1 GFLOPs and 18.1M parameters, surpassing CrossViT-S \cite{chen2021crossvit} at 81.0 with 5.6 GFLOPs and PVT-M at 81.2 with 6.7 GFLOPs. Across all scales, ConvNeur also outperforms pure convolutional baselines such as RepVGG \cite{ding2021repvgg}, from A0 at 72.4 with 1.4 GFLOPs to B3g4 at 80.2 with 16.1 GFLOPs, highlighting the importance of injecting explicit global context rather than deepening or widening a single local stream.
These gains follow the design. The locality-preserving branch maintains translation-friendly priors and fine detail. The global branch aggregates scene-level cues in a compact memory space on chunked tokens, so the cost of global reasoning remains modest. A learned gate then modulates the local stream rather than overwriting it, avoiding feature drift while still exploiting global evidence. Because the memory path is bounded by its bottleneck width and chunk length, its cost grows slowly as models scale, which explains why ConvNeur stays ahead of pooling-based token mixers, frequency-domain globalizers, and hierarchical Transformers that operate global interactions at full width.

\begin{figure*}
    \centering
    \includegraphics[width=\linewidth]{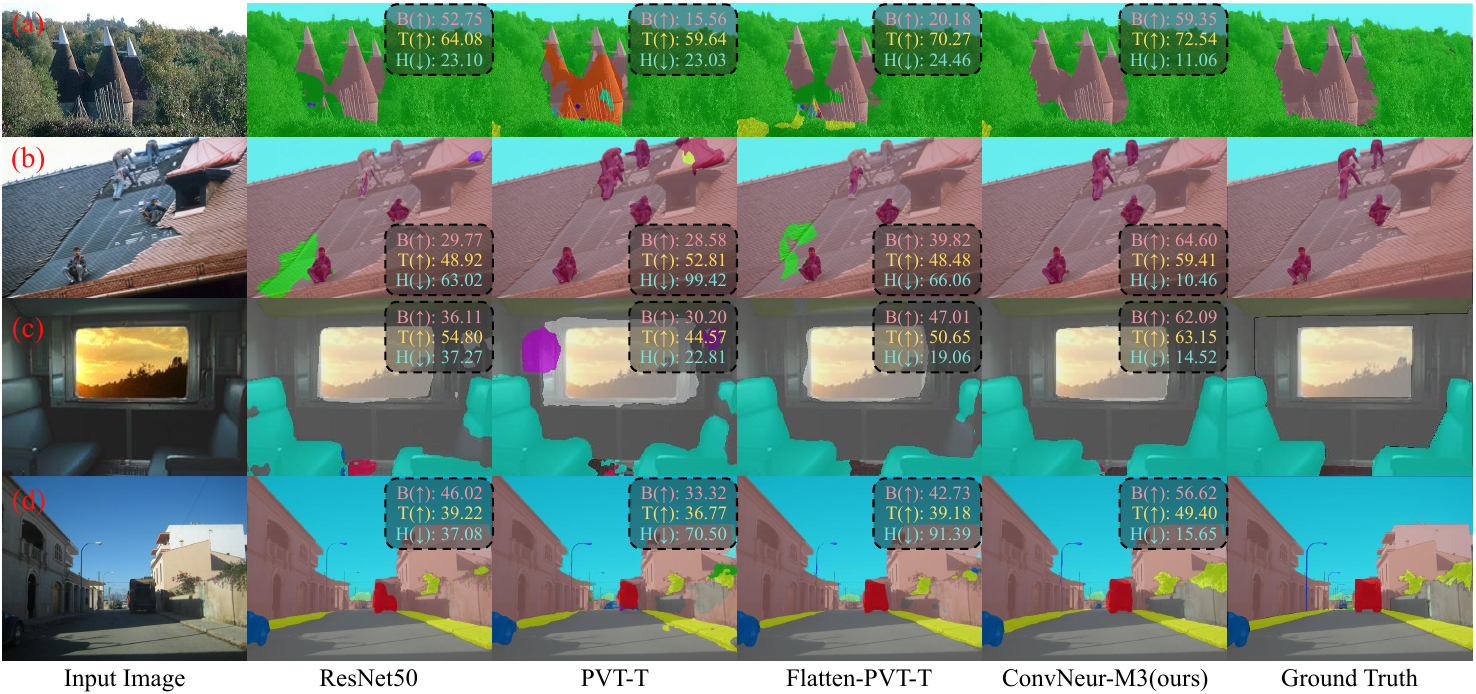}
    \caption{\textbf{Qualitative results on the ADE20K \cite{zhou2017scene} dataset.} All examples are from the validation set. In the figure, \textcolor[RGB]{239,148,158}{B} represents Boundary F1 score, \textcolor[RGB]{252,201,46}{T} represents Trimap-based mIoU, and \textcolor[RGB]{106,216,207}{H} represents Hausdorff distance.}
    \label{fig:vis}
    \vspace{-0.2cm}
\end{figure*}

\subsection{Object Detection}

\textbf{Settings}. We train on COCO 2017 \cite{lin2014microsoft} and report results on the validation set. We adopt Cascade Mask R-CNN \cite{cai2019cascade} in MMDetection \cite{chen2019mmdetection} as the detector and train with the 1x schedule. Images are resized with the shorter side at 800 and the longer side capped at 1333. Training uses 4 NVIDIA RTX 3090 GPUs with 2 images per GPU, giving a total batch size of 8. The 1x schedule runs for 12 epochs with a 500 iteration linear warm-up, followed by multi-step decay at epochs 8 and 11 with gamma 0.1. The optimizer is SGD with a learning rate of 0.02, momentum 0.9, and weight decay 1e-4. All other components follow MMDetection defaults, and both training and inference are single-scale without test-time augmentation.
For the global branch, the higher input resolution in COCO 2017 motivates a larger chunk size of 4096 so that each memory update aggregates more tokens, which stabilizes the surprise-driven updates. We also reduce the inner-loop step size from 1 to 1e-3 to discourage overly frequent memory writes under sparse box and mask supervision.

\noindent\textbf{Results}. As shown in Tab.~\ref{tab:coco}, ConvNeur improves both boxes and masks while keeping the detector unchanged. With the small backbone, ConvNeur-M2 reaches 41.2 AP and 44.5 $\text{AP}_{75}$ for detection, surpassing ResNet-50 \cite{he2016deep} at 38.0 and PVT-T \cite{wang2021pyramid} at 36.7. The gain comes from scene-level cues retrieved by the global memory that help disambiguate scale and context, while the gate preserves edges and small parts so localization tightens rather than blurs. Upder the same setting, M2 attains 36.4 AP mask, stronger than ResNet-50 at 34.4 and PVT-T at 35.1, indicating that modulated global guidance sharpens boundaries in crowded regions.
Scaling the backbone reinforces the trend. ConvNeur-M3 delivers 42.4 AP, exceeding ResNeXt-101-32x4d \cite{xie2017aggregated} at 41.9 and PVTv2-B1 \cite{wang2022pvt} at 41.8. The improvement is consistent with the per-stage placement of memory, which aligns global guidance with the FPN pyramid and provides cross-scale hints without inflating head compute. For masks, M3 reaches 39.0 AP mask, higher than PVTv2-B1 at 38.8, suggesting that chunked memory helps separate overlapping instances and maintain fine contours. Across both scales, ConvNeur shows higher $\text{AP}_{75}$ at comparable $\text{AP}_{50}$, which points to better localization quality rather than only higher recall, matching the design goal of decoupled global-local modeling with learned gating.

\subsection{Semantic Segmentation}

\textbf{Settings}. We train on ADE20K \cite{zhou2017scene} with Semantic FPN \cite{kirillov2019panoptic} and UperNet \cite{xiao2018unified} in MMSegmentation \cite{mmseg2020}. Images are resized so that the long side is 2048 and the short side is 512. During training we apply a random crop of $512^2$. Experiments run on 4 NVIDIA RTX 3090 GPUs with 4 images per GPU for a total batch size of 16. Optimization uses AdamW \cite{loshchilov2017decoupled} with base learning rate 6e-5, betas 0.9 and 0.999, and weight decay 0.01. We also apply gradient clipping with max norm 1.0, and use layer-wise learning rate multipliers of 10 for the neck and decode head. The schedule warms up linearly for 1500 iterations, then follows a polynomial decay until 80k iterations.

\begin{table}[]
\caption{\textbf{Comparison results of semantic segmentation performance on the ADE20K \cite{zhou2017scene} validation set.} The FLOPs are computed with an input image at the resolution of $512\times2048$. In the table, S-FPN is short for the Semantic FPN \cite{kirillov2019panoptic} model.}
\label{tab:seg}
\vspace{-0.2cm}
\setlength{\tabcolsep}{5pt}
\resizebox{\columnwidth}{!}{%
\begin{tabular}{
>{\columncolor[HTML]{ECF4FF}}c
>{\columncolor[HTML]{EFEFEF}}l 
>{\columncolor[HTML]{EFEFEF}}c 
>{\columncolor[HTML]{EFEFEF}}c 
>{\columncolor[HTML]{EFEFEF}}c }
\specialrule{1pt}{0pt}{0pt}
\cellcolor[HTML]{D6D8EB}\textbf{Head} 
& \cellcolor[HTML]{D6D8EB}\textbf{Method} 
& \cellcolor[HTML]{D6D8EB}\textbf{FLOPs(G)} $\downarrow$ 
& \cellcolor[HTML]{D6D8EB}\textbf{mIoU} $\uparrow$ 
& \cellcolor[HTML]{D6D8EB}\textbf{mAcc} $\uparrow$      \\\hline\hline

& ResNet50 \cite{he2016deep}          & 183          & 36.59 & 46.85 \\
& PVT-T \cite{wang2021pyramid}        & 158          & 36.57 & 46.72 \\
& FLatten-PVT-T \cite{han2023flatten} & 169          & 37.21 & 48.95 \\
& \cellcolor[HTML]{D6D8EB}\textbf{ConvNeur-M2(ours)}          
& \cellcolor[HTML]{D6D8EB}\textbf{106} 
& \cellcolor[HTML]{D6D8EB}39.17      
& \cellcolor[HTML]{D6D8EB}49.85 \\
\multirow{-5}{*}{\begin{tabular}[c]{@{}c@{}}S-FPN\\ \cite{kirillov2019panoptic}\end{tabular}}
& \cellcolor[HTML]{D6D8EB}\textbf{ConvNeur-M3(ours)}          
& \cellcolor[HTML]{D6D8EB}123          
& \cellcolor[HTML]{D6D8EB}\textbf{41.42}      
& \cellcolor[HTML]{D6D8EB}\textbf{52.35} \\\hline\hline

& ResNet50 \cite{he2016deep}          & 953          & 42.10 & 52.91 \\
& DeiT-S \cite{touvron2021training}   & 1217         & 43.92 & 54.33 \\
& Swin-T \cite{liu2021swin}           & 945          & 44.51 & 55.61 \\
& FLatten-Swin-T \cite{han2023flatten}& 946          & 44.82 & 57.01 \\
\multirow{-5}{*}{\begin{tabular}[c]{@{}c@{}}UperNet\\ \cite{xiao2018unified}\end{tabular}} 
& \cellcolor[HTML]{D6D8EB}\textbf{ConvNeur-M4(ours)}          
& \cellcolor[HTML]{D6D8EB}\textbf{908} 
& \cellcolor[HTML]{D6D8EB}\textbf{45.72} 
& \cellcolor[HTML]{D6D8EB}\textbf{57.61} \\
\specialrule{1pt}{0pt}{0pt}
\end{tabular}}
\vspace{-0.3cm}
\end{table}

\noindent\textbf{Quantitative Results}. Tab.~\ref{tab:seg} reports ADE20K under identical heads and pipelines. With Semantic FPN, ConvNeur-M2 attains 39.17 mIoU at 106 GFLOPs, outperforming ResNet-50 \cite{he2016deep} at 36.59 with 183 GFLOPs and PVT-T \cite{wang2021pyramid} at 36.57 with 158 GFLOPs, while ConvNeur-M3 pushes to 41.42 at 123 GFLOPs. The gains reflect scene cues retrieved by the global memory that guide the local stream toward complete regions and clean boundaries, which a purely local backbone with S-FPN tends to miss.
With UperNet, ConvNeur-M4 reaches 45.72 mIoU and 57.61 mAcc at 908 GFLOPs, exceeding Swin-T \cite{liu2021swin} at 44.51 with 945 GFLOPs and Flatten-Swin-T \cite{han2023flatten} at 44.82 with 946 GFLOPs, and also surpassing DeiT-S \cite{touvron2021training} at 43.92 with 1217 GFLOPs. Improvements are strongest at tighter budgets, consistent with keeping global reasoning in a compact memory space. The concurrent rise in mAcc indicates fewer class confusions in cluttered scenes, supporting the view that decoupling global reasoning from local representation benefits dense prediction without inflating head compute.

\noindent\textbf{Qualitative Results}. Fig.~\ref{fig:vis} compares ConvNeur with recent backbones on four challenging scenes selected by low mIoU of prior methods, including (a) occlusion where hillside trees partially hide houses, (b) background clutter where colors closely match the target, (c) high illumination contrast area, and (d) cast shadows that erase part of the object semantics. ConvNeur recovers more complete masks and cleaner edges across all panels, proving that the locality-preserving path keeps fine structures intact, and the global memory produces a context map that suppresses distractors and fills in missing parts by borrowing evidence from distant, unoccluded regions. This global-to-local guidance is especially visible on thin facades behind foliage, furniture against similarly colored walls, reflective surfaces, and shadowed object halves. We also annotate three boundary metrics next to each prediction, including Boundary F1 score, Trimap-based mIoU, and Hausdorff distance. ConvNeur improves all three, suggesting that global guidance together with a decoupled global and local design promotes crisp, well-aligned boundaries. These examples show that separating global reasoning from local representation improves instance completeness and boundary fidelity in complex scenes.

\begin{table}[]
\caption{\textbf{Ablation studies on the ImageNet-1K \cite{deng2009imagenet} dataset.} Rows highlighted in \colorbox{customPurple}{purple} indicate the default configuration. The highest value is highlighted in \textbf{bold}.}
\label{tab:ablation}
\vspace{-0.2cm}
\setlength{\tabcolsep}{5pt}
\resizebox{\columnwidth}{!}{%
\begin{tabular}{
>{\columncolor[HTML]{ECF4FF}}c
>{\columncolor[HTML]{EFEFEF}}c 
>{\columncolor[HTML]{EFEFEF}}c 
>{\columncolor[HTML]{EFEFEF}}c 
>{\columncolor[HTML]{EFEFEF}}c }
\specialrule{1pt}{0pt}{0pt}
\cellcolor[HTML]{D6D8EB}\textbf{Experiment}      
& \cellcolor[HTML]{D6D8EB}\textbf{Varient \#} 
& \cellcolor[HTML]{D6D8EB}\textbf{Method}        
& \cellcolor[HTML]{D6D8EB}\textbf{\begin{tabular}[c]{@{}c@{}}FLOPs\\ (G) $\downarrow$\end{tabular}} 
& \cellcolor[HTML]{D6D8EB}\textbf{\begin{tabular}[c]{@{}c@{}}Top-1\\ Acc $\uparrow$\end{tabular}}\\ \hline\hline

\cellcolor[HTML]{EFEFEF}\textit{Component}              & & &            &               \\ \hline
& \ding{182} & Local-only                                 & \textbf{1.4} & 78.2          \\
& \cellcolor[HTML]{D6D8EB}\textbf{-}                      & \cellcolor[HTML]{D6D8EB}\textbf{Per-stage (ours)} 
& \cellcolor[HTML]{D6D8EB}1.8                             & \cellcolor[HTML]{D6D8EB}80.0 \\
\multirow{-3}{*}{\begin{tabular}[c]{@{}c@{}}Global\\ Memory\end{tabular}}
& \ding{183} & Per-layer                                  & 2.3          & \textbf{80.4} \\ \hline

& \ding{184} & CBAM \cite{woo2018cbam}                    & \textbf{1.5} & 78.4          \\
& \ding{185} & Non-local \cite{wang2018non}               & 3.0          & 79.1          \\
& \ding{186} & Self Attention \cite{vaswani2017attention} & 1.8          & 79.8          \\
& \ding{187} & IR-RWKV \cite{jiang2025rwkv}               & 2.0          & 77.6          \\
& \ding{188} & Titans \cite{behrouz2024titans}            & 3.1          & 79.6          \\
\multirow{-6}{*}{\begin{tabular}[c]{@{}c@{}}Global\\ Branch Type\end{tabular}} 
& \cellcolor[HTML]{D6D8EB}\textbf{-}                      & \cellcolor[HTML]{D6D8EB}\textbf{Ours}            
& \cellcolor[HTML]{D6D8EB}1.8                             & \cellcolor[HTML]{D6D8EB}\textbf{80.0}\\ \hline

& \ding{189} & Addition                                   & \textbf{1.8} & 79.6          \\
& \ding{190} & Concatenate                                & 1.9          & 79.5          \\
\multirow{-3}{*}{\begin{tabular}[c]{@{}c@{}}Fusion\\ Mechanism\end{tabular}} 
& \cellcolor[HTML]{D6D8EB}\textbf{-}                      & \cellcolor[HTML]{D6D8EB}\textbf{Gating (ours)}    
& \cellcolor[HTML]{D6D8EB}\textbf{1.8}                    & \cellcolor[HTML]{D6D8EB}\textbf{80.0}\\ \hline\hline

\cellcolor[HTML]{EFEFEF}\textit{Hyperparameter}         & & &            &               \\ \hline
& \ding{172} & 64                                         & \textbf{1.5} & 78.6          \\
& \cellcolor[HTML]{D6D8EB}\textbf{-}                      & \cellcolor[HTML]{D6D8EB}\textbf{128 (ours)} 
& \cellcolor[HTML]{D6D8EB}1.8                             & \cellcolor[HTML]{D6D8EB}80.0 \\
\multirow{-3}{*}{\begin{tabular}[c]{@{}c@{}}Mem Dim\\$C_m$\end{tabular}}   
& \ding{173} & 256                                        & 2.7          & \textbf{80.3} \\ 
\specialrule{1pt}{0pt}{0pt}
\end{tabular}}
\vspace{-0.4cm}
\end{table}

\subsection{Ablation Study}
We ablate on ConvNeur-M3 using ImageNet-1K with the same schedule as in Sec.~\ref{sec:cls}. We analyze the contribution of each core component and rigorously evaluate the model's computational efficiency, with results summarized in Tab.~\ref{tab:ablation}.

\noindent\textbf{Global Memory (\textit{cf.} \ding{182} \ding{183})}.
We compare three settings: a local-only branch (\ding{182}), adding neural memory once at the beginning of each stage, and adding neural memory at every block (\ding{183}). Relative to the local-only baseline, the per-stage memory lifts Top-1 from 78.2 to 80.0 while adding only 0.4 GFLOPs. This confirms the value of decoupling: the local path preserves inductive priors, and the memory supplies image-level cues that a single stream cannot capture under the similar budget. Activating memory at every block reaches 80.4 but requires 2.3 GFLOPs, which is a small gain for a clear increase in cost. The modest improvement aligns with the high redundancy across adjacent blocks within a stage, where global context changes slowly and the gate acts as a slowly varying conditioner. In contrast, stage boundaries introduce larger shifts due to downsampling and width changes, so a single update per stage provides most of the benefit. We therefore adopt the per-stage design by default.

\noindent\textbf{Global Branch Type (\textit{cf.} \ding{184} \ding{185} \ding{186} \ding{187} \ding{188})}.
We next ask what form the global branch should take. For a fair comparison, all variants use the same bottleneck scaffold and the same global dimension at the same insertion points. CBAM \cite{woo2018cbam} is light at 1.5 GFLOPs but reaches only 78.4, a gain of 0.2 over the local-only baseline, which indicates that reweighting alone does not provide persistent global context. Non-local \cite{wang2018non} reaches 79.1 with 3.0 GFLOPs. Its cost stems from building dense affinities across all spatial positions and applying the response matrix, which introduces quadratic compute and heavy memory traffic. Self attention \cite{vaswani2017attention} attains 79.8 at 1.8 GFLOPs, close to our budget, yet its activation memory and attention maps grow quadratically with tokens, so the footprint rises quickly at higher resolutions. The linear RWKV variant \cite{jiang2025rwkv} yields 77.6 at 2.0 GFLOPs. This likely reflects two factors: recurrence imposes a scan order over the spatial sequence and breaks isotropy in 2D, and a single low-dimensional recurrent state is a narrow conduit that tends to forget long-range cues across distant chunks. Our neural memory instead keeps multiple fast weights and retrieves chunk-level associations without a fixed ordering, which aligns better with the chunked setting. A Titans \cite{behrouz2024titans} style memory reaches 79.6 at 3.1 GFLOPs. Our design removes segmented attention and retains neural memory as the sole global mechanism, reducing FLOPs by more than 40\% while slightly improving accuracy to 80.0. Overall, the proposed neural memory offers the most favorable balance under matched scaffolding.

\begin{figure}
    \centering
    \includegraphics[width=1\linewidth]{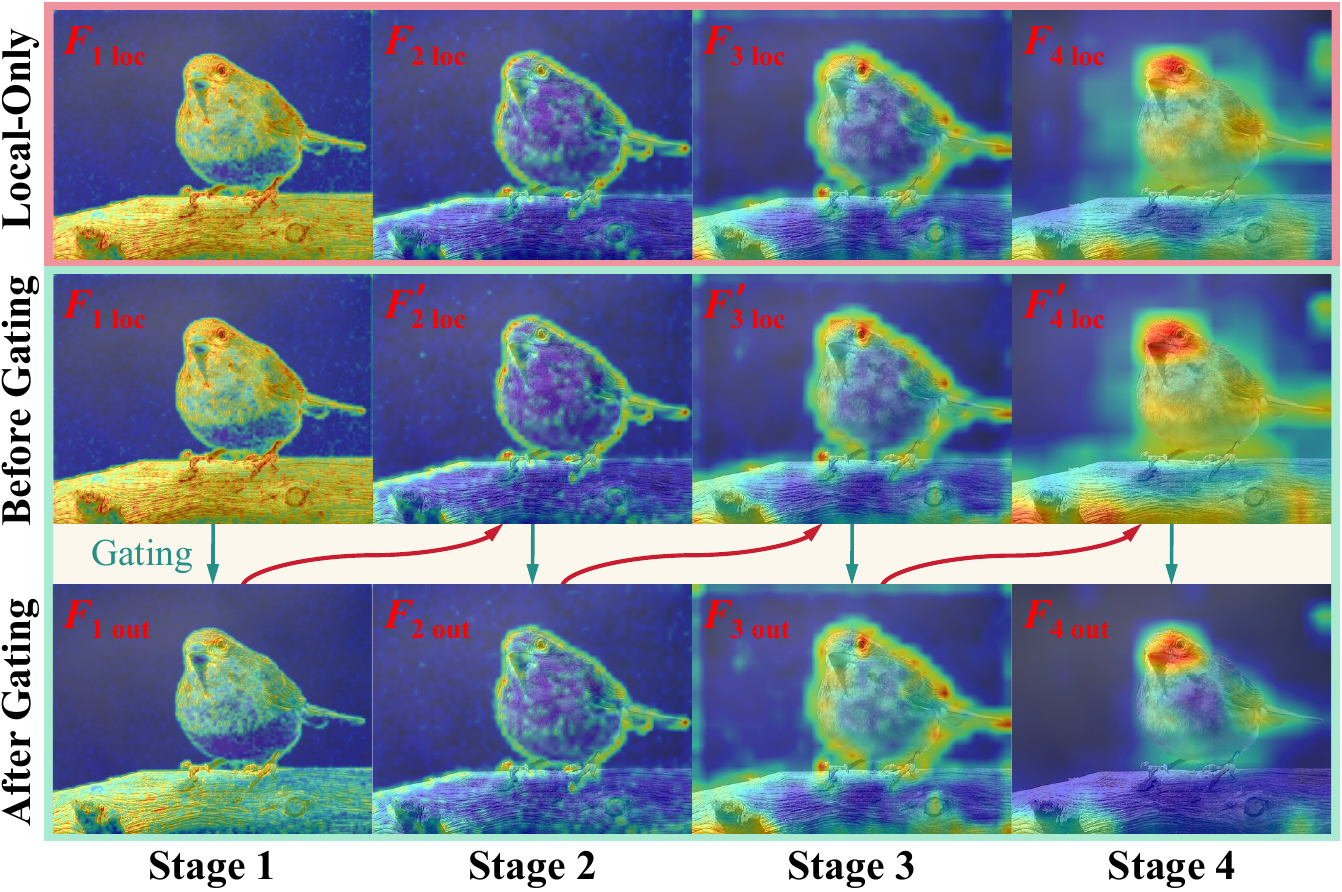}
    \vspace{-0.4cm}
    \caption{\textbf{Stage-wise visualization of global to local modulation.} Columns correspond to the four stages from shallow to deep. Row 1 shows local-only features $F_\text{loc}$ produced by the locality-preserving branch without the global path. Row 2 shows the local features in ConvNeur before gating. Row 3 shows the gated features $F_\text{out}$. \textcolor[RGB]{36,144,135}{Green} arrows mark the gating step.}
    \label{fig:heat}
    \vspace{-0.4cm}
\end{figure}

\noindent\textbf{Fusion Mechanism (\textit{cf.} \ding{189} \ding{190})}.
With the global branch fixed to neural memory, we compare three fusion schemes under matched settings. Simple addition reaches 79.6 at 1.8 GFLOPs. Concatenation reaches 79.5 at 1.9 GFLOPs and introduces 0.7M extra parameters due to the required pointwise reduction. Learned gating attains 80.0 at 1.8 GFLOPs. The small but consistent gain of gating likely comes from its multiplicative calibration, which preserves the residual local path and selectively amplifies or suppresses features per location and per channel. Additive fusion injects global activations directly into the residual stream and can blur fine structures by shifting feature statistics, especially when the global response is strong. Concatenation pays extra compute to learn a similar rescaling after the merge, yet the bottleneck leaves limited headroom for improvement. We therefore adopt gating as the default fusion.

\noindent\textbf{Memory Dimension $C_m$ (\textit{cf.} \ding{172} \ding{173})}.
The global branch runs in a bottlenecked space, so $C_m$ sets its compute share. Raising $C_m$ from 64 to 128 increases FLOPs from 1.5 to 1.8 and lifts Top-1 from 78.6 to 80.0, which is a strong return for a small cost. Pushing to 256 raises FLOPs to 2.7 but reaches only 80.3, showing diminishing gains. Moreover, relative to variant \ding{183}, which deepens the global path by applying memory at every block, simply widening to 256 yields less improvement at higher cost. We set $C_m=128$ as the default since it offers the best balance of accuracy and compute.

\subsection{Mechanism Visualization}
We further probe how the global branch guides the local stream by visualizing three feature views per stage, as shown in Fig.~\ref{fig:heat}, including local-only activations, pre-gate activations in ConvNeur, and the gated result. The influence map is the absolute change between pre-gate and post-gate features, summed over channels and upsampled to the image. The pattern is consistent across stages. The global gate works like a weighted selector that picks the truly informative parts of the local features from a scene-level viewpoint. Early layers show background suppression and cleaner edges. Thin structures and small parts gain contrast while textured clutter fades. Deeper layers become object-centric and highlight semantically relevant regions such as distinctive parts of the target (see $F^\prime_\text{4 loc}\rightarrow F_\text{4 out}$ in Fig.~\ref{fig:heat}). The gate amplifies evidence that supports the current hypothesis and attenuates distractors with similar color or texture.
These observations support the design principle. Global reasoning and local representation are learned on separate paths, and a learned gate converts global cues into pixelwise modulation. This preserves local priors while sharpening boundaries and improving region completeness.

\section{Conclusion}
In this paper, we revisited the long-standing tension between \emph{seeing the whole image} and \emph{keeping local detail}, and argued for a simple design rule: learn global reasoning and local representation separately, then let global cues modulate the local stream. Building on this principle, we introduced \textbf{ConvNeur}, a two-branch architecture with a locality-preserving convolutional path and a compact global path that performs chunked neural memory retrieval and surprise-driven update, followed by learned gating. This decoupling keeps inductive priors intact, makes global aggregation lightweight, and turns global information into targeted guidance rather than a wholesale rewrite.
Experiments on classification, detection, and segmentation show consistent accuracy-efficiency gains under comparable budgets, and ablations confirm that each piece matters: per-stage memory beats local-only baselines, neural memory is a stronger globalizer than common alternatives at matched cost, and gated modulation outperforms naive fusion.
We see ConvNeur as a minimal template for efficient global-local modeling. Future work includes multi-scale and dynamic chunking, extending the memory to video and streaming settings, and refining the theory of the inner fast-weight updates. We hope this perspective encourages architectures that separate roles first and let efficiency follow.

\section*{Acknowledgments}
This work was supported by the National Natural Science Foundation of China (No. 62506169, 62472222), Natural Science Foundation of Jiangsu Province (No. BK20240080), National Defense Science and Technology Industry Bureau Technology Infrastructure Project (JSZL2024606C001).

{
    \small
    \bibliographystyle{ieeenat_fullname}
    \bibliography{main}
}

\end{document}